# Palisade - Prompt Injection Detection Framework

Sahasra Kokkula
*Department of Networking and communication*
*Student of SRM Institute of Science and Technology, Kattankulathur*
Chennai, India
kk4785@srmist.edu.in

Somanathan R
*Department of Networking and communication*
*Student of SRM Institute of Science and Technology, Kattankulathur*
Chennai, India
sr0451@srmist.edu.in

Nandavardhan R
*Department of Networking and communication*
*Student of SRM Institute of Science and Technology, Kattankulathur*
Chennai, India
nr3648@srmist.edu.in

Aashishkumar
*Department of Networking and communication*
*Student of SRM Institute of Science and Technology, Kattankulathur*
Chennai, India
am2567@srmist.edu.in

G Divya
*Department of Networking and communication*
*Faculty of Engineering and Technologoy, SRM Institute of Science and Technology, Kattankulathur*
Chennai, India
divyag2@srmist.edu.in

**Abstract**—Background: The advent of Large Language Models (LLMs) represents a significant milestone in the development of Artificial Intelligence. LLMs have fundamentally altered how machines comprehend and generate human language. However, LLMs remain vulnerable to malicious attacks aimed at manipulating the language model's output. Prompt injection is one such attack where specific inputs (prompts) are designed to manipulate the model's behaviour in unintended or harmful ways. These attacks compromise system integrity, leading to incorrect decisions or outcomes based on the manipulated output.

Conventional detection methodologies typically rely on static or predetermined patterns and rule-based methods to identify injected inputs. While these approaches can be effective against established and straightforward patterns, they often fail to detect more sophisticated threats such as abnormal token sequences and alias substitutions. This limitation of conventional methods results in limited adaptability and high rates of false positives and false negatives.

Method: This paper proposes using NLP techniques to explore and develop an approach for detecting prompt injection with increased accuracy and optimization. For this purpose, a layered approach is utilized, wherein input prompts are screened through three separate layers before they are allowed to reach the LLM, thereby significantly reducing the risk of any injected prompt interacting with the target LLM. This creates a more robust model that can make reliable decisions with minimal to no malicious intervention.

Results: The three layers, Rule based, ML classifier and companion LLM approach, have been tested for accuracy and corresponding confusion matrix was plotted. Among the three layers ML classifier is the most accurate layer.

As we proposed, the multi-layer detection framework that combines the results of each layer to evaluate a final result has a significantly higher accuracy than any one layer in isolation. This increase in accuracy can be attributed to the fact that we will consider a prompt as injected when even one layer reports it as injected, this means we have the least number of false negatives in the combined result. One drawback to flagging a prompt as injected even when one layer reports it as injected is that we have higher number of false positives than the ideal amount, however, in practice, it is preferable to err on the side of caution by flagging prompts as injected, even if some may not be, in order to minimise the risk of overlooking actual injected prompts. While this may result in an increase in false positives, it is more critical to reduce the occurrence of false negatives in this context.

Conclusion:
This research demonstrates a robust multi-layered framework to prompt injection detection, significantly enhancing their security and reliability. By integrating rule-based filtering, a BERT-based machine learning classifier, and a companion LLM, we have achieved a substantial reduction in false negatives, thereby improving the accuracy of prompt injection detection. This methodology not only highlights the vulnerabilities of LLMs but also provides a comprehensive framework for future research and implementation in various applications, ensuring a more secure interaction between humans and AI systems.

*Keywords*—Prompt Injection, NLP, Heuristic-based detection, Large language models.

I. INTRODUCTION

LLM's (large language models) are a revolutionary bridge between humans and machines allowing more natural,

intuitive and versatile human computer interaction. LLMs are used across various fields and industries due to their powerful language understanding and generation capabilities, some of the key fields include customer support, healthcare, financial services, software development, marketing and research-development. Trained on large datasets encompassing multiple languages span across various sources, these models are adept at tasks ranging from simple text completion to complex language analysis. This makes LLMs not just an emerging technology but also a tool that is transfiguring the way we interact with the digital world [1].

A prompt is a text that a user provides to the LLM model and gets back a response. But there might be instances where a threat actor inputs a malicious prompt and the legitimate prompt of the user gets ignored and the malicious instruction will instead be processed by the LLM .This is referred to as a prompt injection attack. Such attacks pose a rising security threat against LLMs like GPT and llama[2]. Various prompt injection techniques exist. One example is direct prompt injection attack, where the attacker has to craft specific inputs that the LLM models consider as legitimate prompts. The attack is executed in real time. Another method used is indirect prompt injection, where there is no direct access to the LLM , but instead the prompt is injected into the sources that the LLM might scrape such as web pages or documents. Stored prompt injection is a subcategory of indirect prompt injection where the malicious instructions are saved within the system model and are executed at a later point of time when the model processes the legitimate user prompt. Prompt injection attacks can lead to reputational damage due to the potential exposure of sensitive information, as well as the risk of misinformation or hateful content being spread. This paper tries to address a method to detect these prompt injection attacks.[3]

his rising prominence of LLMs comes with its own set of vulnerabilities and attacks which include but are not limited to prompt injection attacks, data poisoning attacks, model aversion attacks, adversial attacks, backdoor attacks, evasion attacks, Denial of service(DOS) attacks etc . All these attacks although use different approaches and target different parts of the model have a common goal of disrupting the model performance and reduce the accuracy of its outputs. These attacks on LLMs tend to camouflage themselves in the model, making it difficult to detect and mitigate them before severe damage is done. One such attack is prompt injection where it manipulates the input prompt given to the model to alter its outcome [4].

The first step towards achieving this included choosing a dataset that is comprehensive and large enough to train a robust model. The Hugging Face repository 'deepface/prompt-injections' was chosen due to its relevance to our use case and large dataset size. Following this, the dataset was filtered by language, retaining only English prompts. The dataset was then normalized by removing special characters and standardizing spacing. After normalization, each prompt passes through three filtering layers before reaching the model.
The first layer involves basic rule-based filtering to catch obvious malicious inputs [5]. The second layer uses a lightweight machine learning classifier based on BERT (Bidirectional Encoder Representations from Transformers) [6] to detect patterns of prompt injections. Finally, an LLM companion layer is employed using Groq, a service that provides an API to call the LLM, to detect prompt injections. This multi-layered approach [7] reduces the likelihood of malicious prompts reaching the model, enhancing its security and reliability by preventing attacks that could compromise its functionality.

This paper is organized into several sections to provide a comprehensive overview of the research process and findings. We begin with a review of relevant literature and existing methodologies for prompt injection detection. The subsequent section details the multiple layers of prompt injection detection used to improve accuracy with each subsequent one. These layers include a rule-based approach, an ML classifier using BERT, and a final layer that involves a companion LLM. The results of our experiments, including the effectiveness of the proposed method, are presented and analyzed [8]. Finally, the paper concludes with a discussion of the implications of our findings, limitations of the study, and suggestions for future research.

## II. LITERATURE SURVEY

Prompt injections can present significant security threats to Large Language models (LLMs), allowing malicious actors to manipulate input prompts in order to generate harmful outputs. As LLMs are increasingly embedded in various applications like chatbots and automated systems, the risks associated with such attacks grow substantially. These attacks are categorised into direct prompt injection, where inputs are provided to influence LLM responses directly, and indirect injection, where external systems which feed data into LLMs are manipulated [9].

Most detection mechanisms involve input filtering, although distinguishing between legitimate and malicious prompts is challenging due to the complexity of natural language [10]. One technique suggested is the Signed-Prompt method proposed by Xuchen Suo et al., which uses cryptographic signatures to verify the authenticity of prompts and reduce unauthorised manipulations [11].

Defensive strategies like Reinforcement Learning with Human Feedback (RLHF) have been used to align model behaviour, but they still fall short against sophisticated prompt in In conclusion, while basic methods for detecting and defending against prompt injection exist, they remain limited in the face of evolving attacks. Research, like the development of Signed-Prompt, RLHF highlights the ongoing effort to create more robust defences, though challenges remain in balancing security with model performance [10] [11][12].

All the above mentioned implementations require modification of the target LLM or the prompt, our proposed solution does not require changes to the prompt or the LLM, instead our approach acts as an intermediary between the LLM and the user giving the prompt, detecting if the incoming prompts are injected or not.

## III. METHODOLOGY

### A. Rulebased approach

The dataset utilized for this approach was obtained from the hugging face repository 'deepface/prompt-injections'. It comprises 546 distinct prompts. Each prompt has a binary label associated with it. A label value of 1 denotes that the statement contains prompt injections, while a label of 0 signifies that the prompt is free from such injections. The dataset is then split into test and train portions.

The Natural Language Processing (NLP) component of this approach was implemented using spaCy[13], an open-source library designed for processing and analyzing large quantities of textual data. Although the dataset includes prompts in multiple languages, only English-language prompts are used. This is done to enhance the accuracy of prompt injection detection. While spaCy supports several languages, it may perform optimally with English due to its extensive English-specific resources. Restricting the prompts to English alone ensures that spaCy library's capabilities are fully leveraged for accurate analysis. Different languages have varied grammatical structures and rules that require additional preprocessing, linguistic expertise and potential computing complexities. Another factor in removing other language prompt from the dataset is the prevalence of English in NLP research which means that more resources, reference studies and established benchmarks available

Restricting the dataset to English prompts alone is done using Google translate API. The English prompts are detected and appended to a new csv file.
The prompt sentence is normalized by removing special characters and replacing any multiple spaces with a single space. Then in order to formulate a heuristics-based approach we consider only those prompts that are injected to build rules.

SpaCy's similarity function was used to determine the degree to which a word is similar to a target word. This was accomplished by finding the word embedding using a word to vector[14] conversion algorithm. Then similarity is established between the vectors in a geometrical fashion using techniques such as cosine similarity. This function returns a similarity score ranging from 0 to 1, but we can incorporate a threshold value below which the word is not considered similar.

Nouns, verbs and adjectives are extracted from the injected portion of the dataset then aggregated into a list. To find Nouns, verbs and adjectives which might indicate prompt injection, we start with seed nouns, verbs and adjectives which are most commonly used in prompt injection. These words are passed as the target words to our filter_similar_words this yields 3 lists for injected verbs, nouns and adjectives.

The nouns, verbs and adjectives are classified as injected using the filter_similar_words function which takes in target words which are compared to the aggregated list. This function takes a list of words, a target word, and a threshold value as parameters. The function calculates similarity scores[15] between each word in the aggregated list and the target words with a similarity score above a specified threshold are included.

A scoring function is applied to determine whether a prompt contains an injection or not. This function takes the prompt as input and the previously aggregated list of injected nouns, verbs and adjectives[16]. The input prompt is separated into individual words, each of these words is cross-referenced with the aggregated list to find the index and based on this index, distance is calculated between noun, verb and adjective. This distance was used to find the score on how probable it is the prompt has been injected.

*B. ML Classifier*

This approach leverages the BERT [17] (Bidirectional Encoder Representations from Transformers) package to detect prompt injections. Transformers are a deep learning library developed by Hugging Face, which provides a framework of pre-trained models for NLP tasks.

Although the traditional approach to NLP data ingestion is vectorization, it is not suited for this case, as vectorization assigns a vector of length N to each word or sub-word, which means there is a fixed-size vocabulary[18]. While this approach may work well for conventional NLP tasks, the detection of prompt injections requires a different approach. We must account for the possibility that the prompts may contain abnormal tokens and alias substitutions where a fixed-size vocabulary would not accommodate all possible combinations.

BERT is a transformer-based model used in evaluating contextual word relationships within sentences. The BertTokenizer is used to tokenize the textual data into a specific format that is compatible with the BERT model [19]. This involves converting the text into token IDs, inserting special tokens at the beginning and end of the input, and padding or truncating sentences to ensure uniform input length, aiding in batch processing.

The dataset is divided into training and testing portions using the train_test_split function from scikit-learn. This ensures that the model is trained on one part of the data and validated on another part of data, preventing overfitting. Our approach used the traditional 80/20 split to train and test data respectively.

Although a heuristic approach can detect patterns, it is reliant on fixed patterns, whereas a BERT model can detect patterns from training data, allowing it to identify subtle and sophisticated prompt injections.

Once the data is tokenized, it is structured into a custom PyTorch dataset to allow seamless interaction with the PyTorch Trainer API, which is used for training and evaluation. The model used is the pre-trained BertForSequenceClassification model, which has been customized for text classification applications. Along with the fundamental BERT architecture, this model has an additional linear layer for binary classification [20].

To initialize the model, pre-trained weights from bert-base-uncased were used. The model was trained for two epochs with a batch size of four and gradient accumulation across two steps to increase the batch size.

The weight decay was set to 0.01 to prevent overfitting. In order to accelerate training, mixed-precision training was used. The Hugging Face Trainer API was utilized to streamline the evaluation and training loop. After training, the model was assessed using the test set, and performance was measured using metrics like accuracy and loss.

### C. Companion LLM

The third layer of the multi-layered approach involves a companion LLM. This layer comprises the system prompt, user prompt and a LLM [21].
System prompt is the initial input instructions provided to the LLM on how it should act and respond. For our purposes we have instructed the LLM to be a security detection system which validates user prompts and determines if the user prompt is injected or not. Furthermore, the system prompt is specifically instructed to return a JSON response indicating if the user prompt is injected or not.
User prompt is the input prompt supplied by the user which may or may not contain injection. This is considered as input for the LLM.
In some instances, user prompts can be persuasive enough to bypass the system prompt, causing the LLM to follow the instructions from the user prompt instead, in such cases we have implemented hard-coded error handling which returns "injected" by default.

### D. Complete Layered Evaluation

Once the prompt is evaluated across all layers, the results are aggregated, and a logical OR operator is applied. This means that if any single layer identifies the prompt as injected, we conclude that the prompt is injected, regardless of the results from other layers.

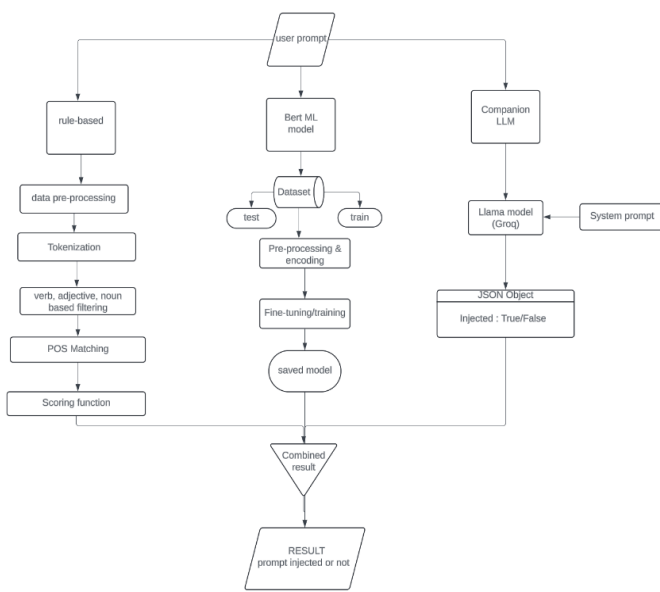

Fig. 1. Multi-Layered Prompt Injection Detection Framework

## IV. RESULTS

To visualize the outcome of this approach of prompt injection detection, a confusion matrix is employed. This matrix compares the actual label with the predicted score to evaluate the performance of the algorithm based on the relationship between the actual and predicted outcomes [22].

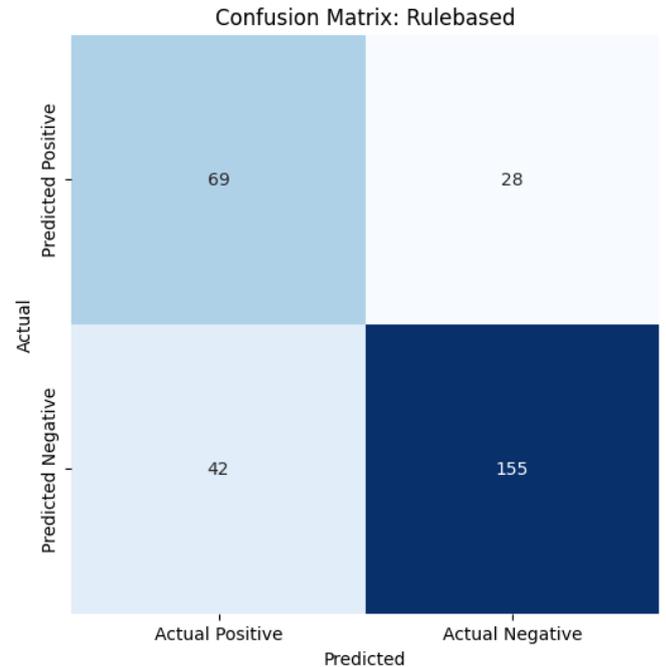

Fig. 2. Confusion matrix for rulebased approach

Figure 2 is the confusion matrix for Rulebased approach which uses heuristics derived from a dataset, This explains the higher than anticipated false negatives

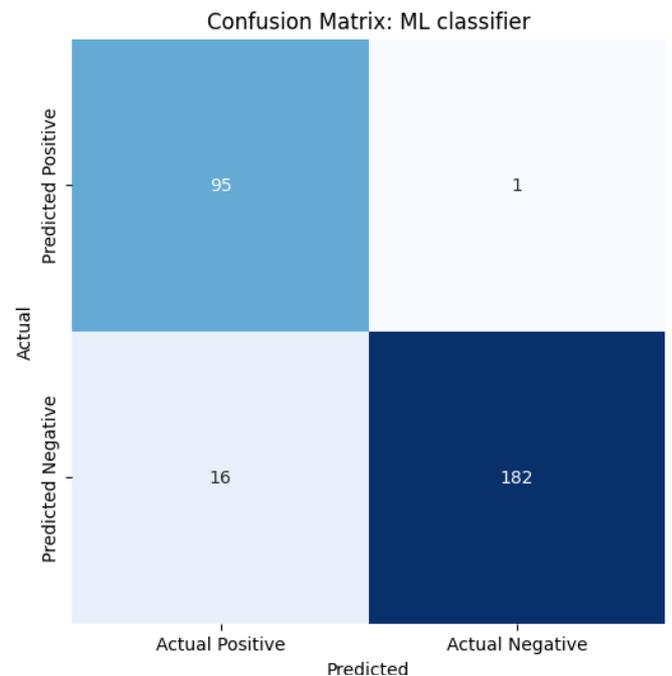

Fig. 3. Confusion matrix for ML BERT approach

Figure 3 represents the confusion matrix for ML based approach which learns from the dataset meaning it can detect patterns which were not present in the dataset verbatim, which increases the accuracy as compared to rulebased approach.

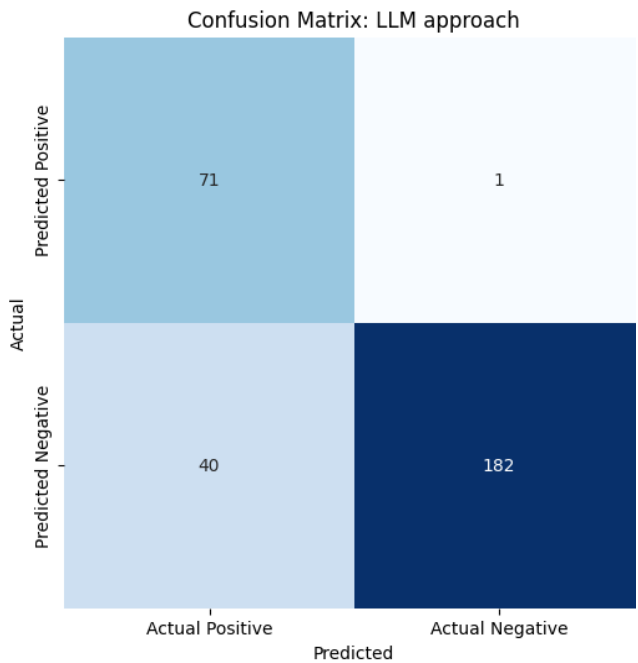

Fig. 4. Confusion matrix for LLM approach

Figure 4 is the confusion matrix for LLM approach which uses a companion LLM to detect if the prompt is injected or not.

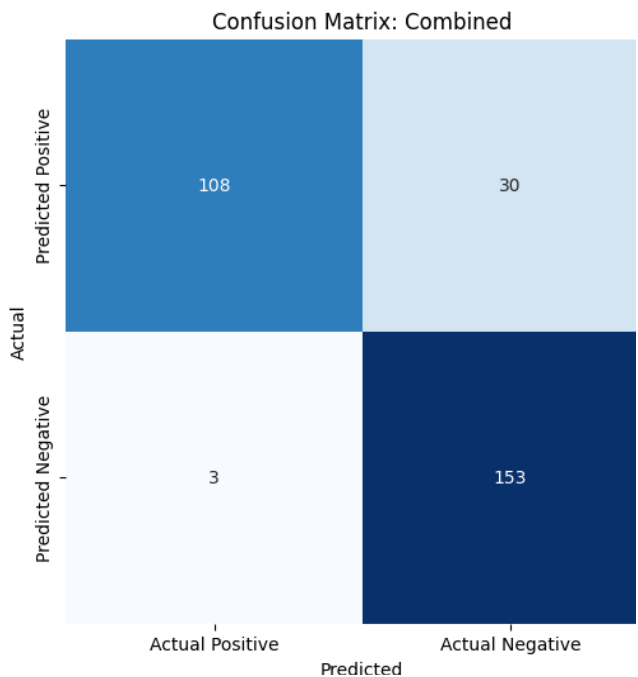

Fig. 5. Confusion matrix for Palisade framework

Figure 5 depicts the confusion matrix for combined result which applies logical OR on the results from all the layers, which explains the exceptionally low false negative. Hence achieving our goal of reduced inaccuracy using our multilayered palisade framework.

## V. CONCLUSION

In this paper, we presented a comprehensive solution on how to handle prompt injection attacks on large language models. We initially highlighted the prevalence of LLMs in this industry and how they are vulnerable to various attacks, with prompt injection attacks being the centerpiece of this research.

We proposed a multilayered framework called **Palisade** that integrates the rule-based approach, a BERT-based machine learning classifier, and a companion LLM. Utilizing the Hugging Face repository with a carefully curated dataset, this framework proposes a strong mechanism to evaluate the prompts given to minimize false positives and negatives.
The results from our experiments indicate that the proposed multilayered Palisade framework significantly improves the effectiveness of prompt injection detection.

Future work should focus on refining the detection methods further and exploring additional layers of security to create a more resilient framework against prompt injection attacks. Overall, our findings contribute valuable insights to the field of natural language processing and security, paving the way for the development of more secure LLMs capable of operating in environments susceptible to malicious intervention.